\documentclass[10pt,twocolumn,letterpaper]{article}

\usepackage{cvpr}
\usepackage{times}
\usepackage{epsfig}
\usepackage{graphicx}
\usepackage{subcaption}
\usepackage{amsmath}
\usepackage{amssymb}
\usepackage[font={small}]{caption}
\usepackage{multirow}
\usepackage{diagbox}
\usepackage{ifthen}
\usepackage{bm}
\newcommand{\redbf}[1]{{\textbf{\color{red}{#1}}}} 
\newcommand{\blueud}[1]{{\underline{\color{blue}{#1}}}} 
\usepackage{pifont}
\newcommand{\cmark}{\ding{51}}%
\newcommand{\xmark}{\ding{55}}%
\usepackage[hang,flushmargin]{footmisc} 
\usepackage[pagebackref=true,breaklinks=true,letterpaper=true,colorlinks,bookmarks=false]{hyperref}
\cvprfinalcopy 
\usepackage{cleveref}

\def\cvprPaperID{59} 

\begin{document}

\title{\vspace{-1cm}EDVR: Video Restoration with Enhanced Deformable Convolutional Networks}
\author{
	Xintao Wang$^{1}$ \hspace{10pt} Kelvin C.K. Chan$^{2}$ \hspace{10pt} Ke Yu$^{1}$ \hspace{10pt} Chao Dong$^{3}$
	\hspace{10pt}Chen Change Loy$^{2}$\\
	\vspace{-0.15cm}
	\small{$^{1}$CUHK - SenseTime Joint Lab, The Chinese University of Hong Kong} 
	\hspace{5pt}
	\small{$^{2}$Nanyang Technological University, Singapore}\\
	\vspace{-0.15cm}
	\small{$^{3}$SIAT-SenseTime Joint Lab, Shenzhen Institutes of Advanced Technology, Chinese Academy of Sciences} \\
	{\tt\small \{wx016, yk017\}@ie.cuhk.edu.hk \hspace{3pt} chao.dong@siat.ac.cn\hspace{3pt} \{chan0899, ccloy\}@ntu.edu.sg}\\
	\vspace{-0.7cm}
}

\maketitle

\vspace{-0.4cm}
\begin{abstract}
\vspace{-0.2cm}
Video restoration tasks, including super-resolution, deblurring, etc, are drawing increasing attention in the computer vision community. A challenging benchmark named REDS is released in the NTIRE19 Challenge. This new benchmark challenges existing methods from two aspects: (1) how to align multiple frames given large motions, and (2) how to effectively fuse different frames with diverse motion and blur. 
In this work, we propose a novel Video Restoration framework with Enhanced Deformable convolutions, termed EDVR, to address these challenges. First, to handle large motions, we devise a Pyramid, Cascading and Deformable (PCD) alignment module, in which frame alignment is done at the feature level using deformable convolutions in a coarse-to-fine manner. Second, we propose a Temporal and Spatial Attention (TSA) fusion module, in which attention is applied both temporally and spatially, so as to emphasize important features for subsequent restoration. 
Thanks to these modules, our EDVR wins the champions and outperforms the second place by a large margin in all four tracks in the NTIRE19 video restoration and enhancement challenges. EDVR also demonstrates superior performance to state-of-the-art published methods on video super-resolution and deblurring.
The code is available at \url{https://github.com/xinntao/EDVR}.
\end{abstract}
\vspace{-0.5cm}


\section{Introduction}

In this paper, we describe our winning solution in the NTIRE 2019 challenges on video restoration and enhancement.
The challenge releases a valuable benchmark, known as REalistic and Diverse Scenes dataset (REDS)~\cite{Nah_2019_CVPR_Workshops_REDS}, for the aforementioned tasks. In comparison to existing datasets, videos in REDS contain larger and more complex motions, making it more realistic and challenging. The competition enables fair comparisons among different algorithms and promotes the progress of video restoration.

Image restoration tasks such as super-resolution (SR)~\cite{dong2014learning,lim2017enhanced,timofte2017ntire,ledig2017photo,wang2018recovering,zhang2018image} and deblurring~\cite{nah2017deep,kupyn2017deblurgan,tao2018scale} have experienced significant improvements over the last few years thanks to deep learning.
The successes encourage the community to further attempt deep learning on the more challenging video restoration problems.
Earlier studies~\cite{takeda2009super,dai2015dictionary,shahar2011space,liao2015video,kappeler2016video} treat video restoration as a simple extension of image restoration.
The temporal redundancy among neighboring frames is not fully exploited.
Recent studies~\cite{caballero2017real,xue2017video,tao2017detail,sajjadi2018frame} address the aforementioned problem with more elaborated pipelines that typically consist of four components, namely feature extraction, alignment, fusion, and reconstruction.
The challenge lies in the design of the alignment and fusion modules when a video contains occlusion, large motion,  and severe blurring. To obtain high-quality outputs, one has to (1) align and establish accurate correspondences among multiple frames, and (2) effectively fuse the aligned features for reconstruction.

\begin{figure}[!t]
    \begin{center}
        \includegraphics[width=\linewidth]{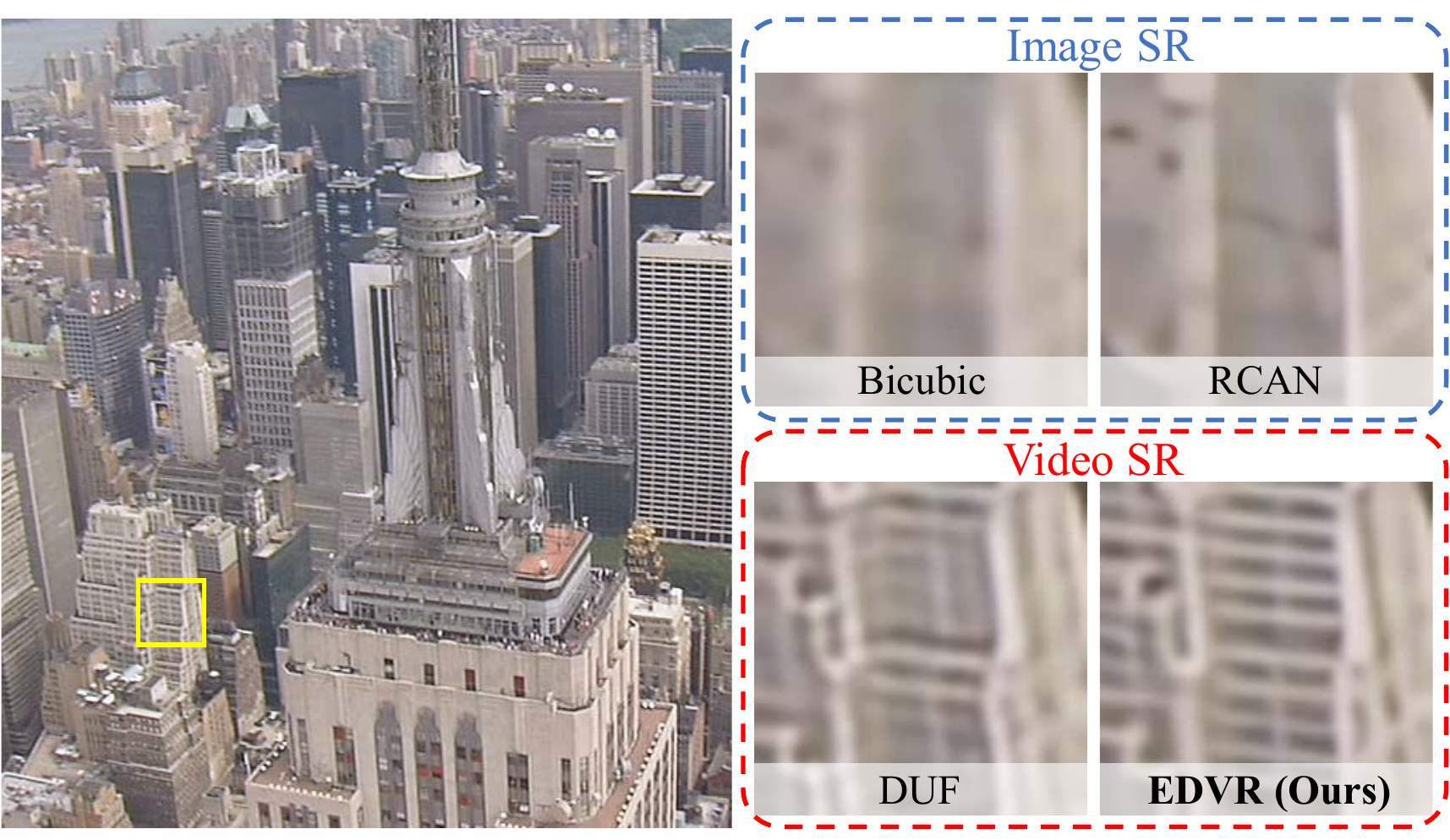}
        \vspace{-0.7cm}
        \caption{A comparison between image super-resolution and video super-resolution ($\times4$). RCAN~\cite{zhang2018image} and DUF~\cite{jo2018deep} are the state-of-the-art methods of image and video super-resolution, respectively.}
        \label{fig:teaser}
        \vspace{-0.8cm}
    \end{center}
\end{figure}

\noindent\textbf{Alignment.}
Most existing approaches perform alignment by explicitly estimating optical flow field between the reference and its neighboring frames~\cite{caballero2017real,xue2017video,kim2018spatio}. The neighboring frames are warped based on the estimated motion fields. Another branch of studies achieve implicit motion compensation by dynamic filtering~\cite{jo2018deep} or deformable convolution~\cite{tian2018tdan}.
REDS imposes a great challenge to existing alignment algorithms.
In particular, precise flow estimation and accurate warping can be challenging and time-consuming for flow-based methods. In the case of large motions, it is difficult to perform motion compensation either explicitly or implicitly within a single scale of resolution.

\noindent\textbf{Fusion.}
Fusing features from aligned frames is another critical step in the video restoration task. Most existing methods either use convolutions to perform early fusion on all frames~\cite{caballero2017real} or adopt recurrent networks to gradually fuse multiple frames~\cite{sajjadi2018frame, haris2019recurrent}. 
Liu~\etal~\cite{liu2017robust} propose a temporal adaptive network that can dynamically fuse across different
temporal scales. None of these existing methods consider the underlying visual informativeness on each frame -- different frames and locations are not equally informative or beneficial to the reconstruction, as some frames or regions are affected by imperfect alignment and blurring.

\noindent\textbf{Our Solution.}
We propose a unified framework, called EDVR, which is extensible to various video restoration tasks, including super-resolution and deblurring. 
The cores of EDVR are (1) an alignment module known as Pyramid, Cascading and Deformable convolutions (PCD), and (2) a fusion module known as Temporal and Spatial Attention (TSA).

The PCD module is inspired by TDAN~\cite{tian2018tdan} in using deformable convolutions to align each neighboring frame to the reference frame at the feature level. 
Different from TDAN, we perform alignment in a coarse-to-fine manner to handle large and complex motions.
Specifically, we use a pyramid structure that first aligns features in lower scales with coarse estimations, and then propagates the offsets and aligned features to higher scales to facilitate precise motion compensation, similar to the notion adopted in optical flow estimation~\cite{hui18liteflownet,ilg2017flownet}. Moreover, we cascade an additional deformable convolution after the pyramidal alignment operation to further improve the robustness of alignment.

The proposed TSA is a fusion module that helps aggregate information across multiple aligned features. 
To better consider the visual informativeness on each frame, we introduce temporal attention by computing the element-wise correlation between the features of the reference frame and each neighboring frame. 
The correlation coefficients then weigh each neighboring feature at each location, indicating how informative it is for reconstructing the reference image. The weighted features from all frames are then convolved and fused together. 
After the fusion with temporal attention, we further apply spatial attention to assign weights to each location in each channel to exploit cross-channel and spatial information more effectively.

We participated in all the four tracks in the video restoration and enhancement challenges~\cite{Nah_2019_CVPR_Workshops_SR,Nah_2019_CVPR_Workshops_Deblur}, including video super-resolution (clean/blur) and video deblurring (clean/compression artifacts). 
Thanks to the effective alignment and fusion modules, our EDVR has won the champion in all the four challenging tracks, demonstrating the effectiveness and the generalizability of our method. In addition to the competition results, we also report comparative results on existing benchmarks of video super-resolution and deblurring. Our EDVR shows superior performance to state-of-the-art methods in these video restoration tasks.

\begin{figure*}[!t]
	\vspace{-0.4cm}
	\begin{center}
		\includegraphics[width=\linewidth]{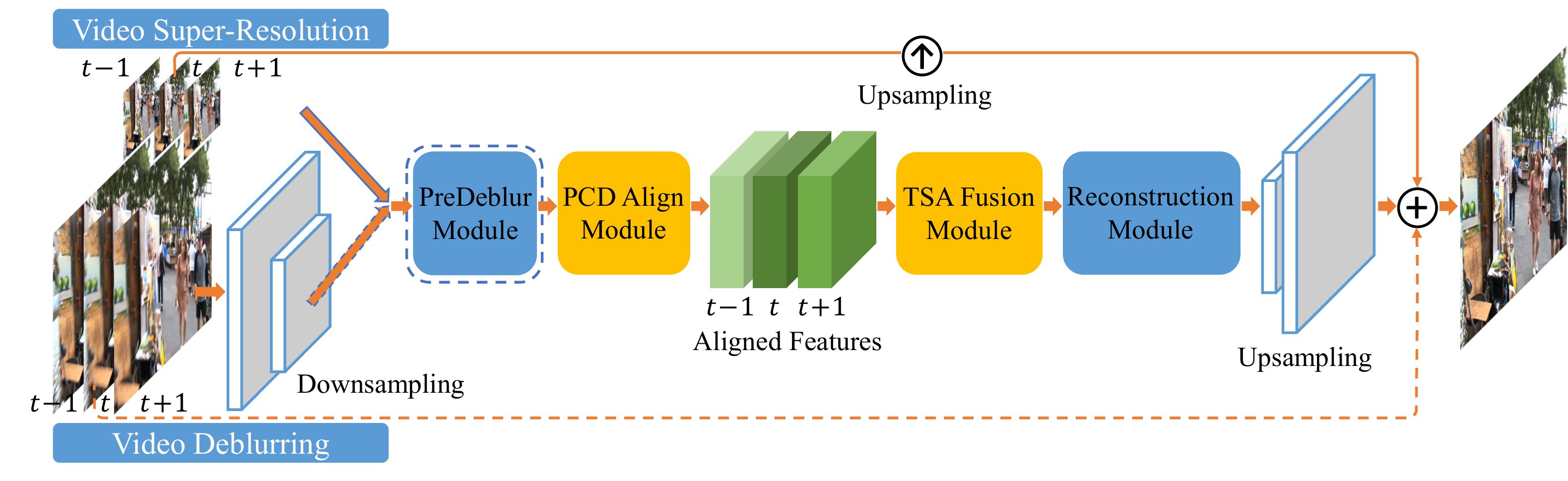}
		\vspace{-0.7cm}
		\caption{\textbf{The EDVR framework}. It is a unified framework suitable for various video restoration tasks, \eg, super-resolution and deblurring. Inputs with high spatial resolution are first down-sampled to reduce computational cost. Given blurry inputs, a PreDeblur Module is inserted before the PCD Align Module to improve alignment accuracy. We use three input frames as an illustrative example.}
		\label{fig:overall_structure}
		\vspace{-0.65cm}
	\end{center}
\end{figure*}


\section{Related Work}
\noindent\textbf{Video Restoration.}
Since the pioneer work of SRCNN~\cite{dong2014learning}, deep learning methods have brought significant improvements in image and video super-resolution~\cite{lim2017enhanced,timofte2017ntire,ledig2017photo,liu2018non,yu2018crafting,yu2019path,caballero2017real,liu2017robust,sajjadi2018frame,tao2017detail,xue2017video}. For video super-resolution, temporal alignment plays an important role and has been extensively studied. Several methods~\cite{caballero2017real,tao2017detail,sajjadi2018frame} use optical flow to estimate the motions between images and perform warping.
However, accurate flow is difficult to obtain given occlusion and large motions. TOFlow~\cite{xue2017video} also reveals that the standard optical flow is not the optimal motion representation for video restoration. 
DUF~\cite{jo2018deep} and TDAN~\cite{tian2018tdan} circumvent the problem by implicit motion compensation and surpass the flow-based methods. Our EDVR also enjoys the merits of implicit alignment, with a pyramid and cascading architecture to handle large motions.

Video deblurring also benefits from the development of learning-based methods~\cite{kim2018dynamic,ma2015handling,pan2017simultaneous,su2017deep}. Several approaches~\cite{su2017deep,zhang2019adversarial} directly fuse multiple frames without explicit temporal alignment, because the existence of blur increases the difficulty of motion estimation. Unlike these approaches, we attempt to acquire information from multiple frames using alignment, with a slight modification that an image deblurring module is added prior to alignment when there is a blur.

\noindent\textbf{Deformable Convolution.} 
Dai~\etal~\cite{dai2017deformable} first propose deformable convolutions, in which additional offsets are learned to allow the network to obtain information away from its regular local neighborhood, improving the capability of regular convolutions.
Deformable convolutions are widely used in various tasks such as video object detection~\cite{bertasius2018object}, action recognition~\cite{zhao2018trajectory}, semantic segmentation~\cite{dai2017deformable}, and video super-resolution~\cite{tian2018tdan}. In particular, TDAN~\cite{tian2018tdan} uses deformable convolutions to align the input frames at the feature level without explicit motion estimation or image warping. Inspired by TDAN, our PCD module adopts deformable convolution as a basic operation for alignment.

\noindent\textbf{Attention Mechanism.}
Attention has proven its effectiveness in many tasks~\cite{vaswani2017attention,woo2018cbam,liu2017robust,liu2018non,zhang2018image}. For example, in video SR, Liu~\etal~\cite{liu2017robust} learn a set of weight maps to weigh the features from different temporal branches. Non-local operations~\cite{wang2018non} compute the response at a position as a weighted sum of the features at all positions for capturing long-range dependencies.
Motivated by the success of these works, we employ both temporal and spatial attention in our TSA fusion module to allow different emphases on different temporal and spatial locations.

\vspace{-0.1cm}
\section{Methodology}
\vspace{-0.1cm}
\subsection{Overview}
\vspace{-0.1cm}
\noindent Given $2N{+}1$ consecutive low-quality frames $I_{[t-N:t+N]}$, we denote the middle frame $I_t$ as the reference frame and the other frames as neighboring frames. The aim of video restoration is to estimate a high-quality reference frame $\hat{O}_t$, which is close to the ground truth frame $O_t$.
The overall framework of the proposed EDVR is shown in Fig.~\ref{fig:overall_structure}.
It is a generic architecture suitable for several video restoration tasks, including super-resolution, deblurring, denoising, de-blocking, \etc.

Take video SR as an example, EDVR takes $2N{+}1$ low-resolution frames as inputs and generates a high-resolution output.
Each neighboring frame is aligned to the reference one by the PCD alignment module at the feature level.
The TSA fusion module fuses image information of different frames.
The details of these two modules are described in Sec.~\ref{subsec:pcd} and Sec.~\ref{subsec:tsa}.
The fused features then pass through a reconstruction module, which is a cascade of residual blocks in EDVR and can be  easily replaced by any other advanced modules in single image SR~\cite{wang2018esrgan,zhang2018image}.
The upsampling operation is performed at the end of the network to increase the spatial size.
Finally, the high-resolution frame $\hat{O}_t$ is obtained by adding the predicted image residual to a direct upsampled image. 

For other tasks with high spatial resolution inputs, such as video deblurring, the input frames are first downsampled with strided convolution layers. Then most computation is done in the low-resolution space, which largely saves the computational cost. The upsampling layer at the end will resize the features back to the original input resolution.
A PreDeblur module is used before the alignment module to pre-process blurry inputs and improve alignment accuracy.

Though a single EDVR model could achieve state-of-the-art performance, we adopt a two-stage strategy to further boost the performance in NTIRE19 competition. Specifically, we cascade the same EDVR network but with shallower depth to refine the output frames of the first stage.
The cascaded network can further remove the severe motion blur that cannot be handled by the preceding model.
The details are presented in Sec.~\ref{subsec:twostage}.

\subsection{Alignment with Pyramid, Cascading and\\Deformable Convolution}
\label{subsec:pcd}

We first briefly review the use of deformable convolution for alignment~\cite{tian2018tdan}, \ie, aligning features of each neighboring frame to that of the reference one.
Different from optical-flow based methods, deformable alignment is applied on features of each frame, denoted by $F_{t+i}, i {\in} [{-}N{:}{+}N]$.
We use the modulated deformable module~\cite{zhu2018deformable}. Given a deformable convolution kernel of $K$ sampling locations, we denote $w_k$ and $\mathbf{p}_k$ as the weight and the pre-specified offsets for the $k$-th location, respectively. For instance, a $3{\times}3$ kernel is defined with $K{=}9$ and $\mathbf{p}_k{\in}\{(-1,-1),(-1,0),\cdots,(1,1)\}$.
The aligned features ${F_{t+i}^a}$ at each position $\mathbf{p}_0$ can then be obtained by:
\vspace{-0.2cm}
\begin{equation}\label{equ:mdcn}
{F_{t+i}^a}(\mathbf{p}_0)=\sum_{k=1}^{K} w_k\cdot F_{t+i}(\mathbf{p}_0+\mathbf{p}_k+\Delta \mathbf{p}_k)\cdot \Delta m_k.
\vspace{-0.1cm}
\end{equation}
The learnable offset $\Delta \mathbf{p}_k$ and the modulation scalar $\Delta m_k$ are predicted from concatenated features of a neighboring frame and the reference one: 
\vspace{-0.1cm}
\begin{equation}
\Delta \mathbf{P}_{t+i}=f(\ [F_{t+i}, F_t]\ ),\quad i \in [{-}N:{+}N]
\vspace{-0.1cm}
\end{equation}
where $\Delta \mathbf{P}{=}\{\Delta \mathbf{p}\}$, $f$ is a general function consisting several convolution layers, and $[\cdot, \cdot]$ denotes the concatenation operation. For simplicity, we only consider learnable offsets $\Delta \mathbf{p}_k$ and ignore modulation $\Delta m_k$ in the descriptions and figures. 
As $\mathbf{p}_0+\mathbf{p}_k+\Delta \mathbf{p}_k$ is fractional, bilinear interpolation is applied as in~\cite{dai2017deformable}.

\begin{figure}[!t]
	\vspace{-0.5cm}
	\begin{center}
		\includegraphics[width=0.9\linewidth]{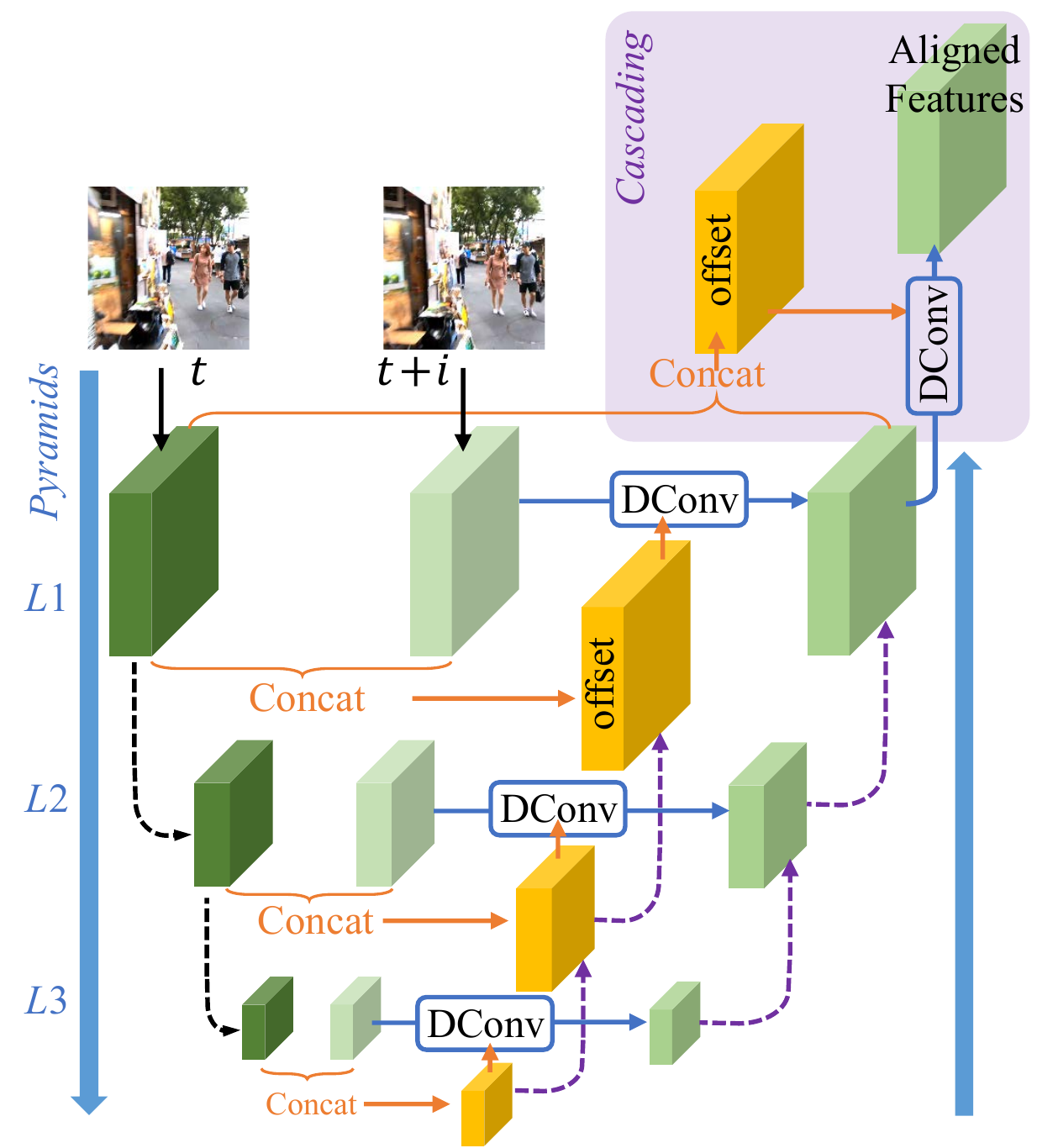}
		\vspace{-0.3cm}
		\caption{PCD alignment module with Pyramid, Cascading and Deformable convolution.}
		\label{fig:pcd_align}
		\vspace{-0.6cm}
	\end{center}
\end{figure}

To address complex motions and large parallax problems in alignment, we propose PCD module based on well-established principles in optical flow: pyramidal processing~\cite{ranjan2016optical,sun2018pwc} and cascading refinement~\cite{hui18liteflownet,hui2019lightweight,ilg2017flownet}.
Specifically, as shown with black dash lines in Fig.~\ref{fig:pcd_align}, to generate feature $F_{t+i}^l$ at the $l$-th level, we use strided convolution filters to downsample the features at the $(l{-}1)$-th pyramid level by a factor of 2, obtaining $L$-level pyramids of feature representation.
At the $l$-th level, offsets and aligned features are predicted also with the ${\times}2$ upsampled offsets and aligned features from the upper $(l{+}1)$-th level, respectively (purple dash lines in Fig.~\ref{fig:pcd_align}):
\vspace{-0.3cm}
\begin{equation}
\Delta \mathbf{P}_{t+i}^l=f(\ [F_{t+i}, F_{t}],\ (\Delta \mathbf{P}_{t+i}^{l+1})^{\uparrow 2}\ ),
\vspace{-0.1cm}
\end{equation}
\vspace{-0.5cm}
\begin{equation}
{(F_{t+i}^a)}^{l}=g(\ \text{DConv}(F_{t+i}^l,\Delta \mathbf{P}_{t+i}^l),\ ({(F_{t+i}^a)}^{l+1})^{\uparrow 2}\ ),
\end{equation}
where $(\cdot)^{\uparrow s}$ refers to upscaling by a factor $s$, DConv is the deformable convolution described in Eqn.~\ref{equ:mdcn}, and $g$ is a general function with several convolution layers. Bilinear interpolation is adopted to implement the ${\times}2$ upsampling. We use three-level pyramid, \ie, $L{=}3$, in EDVR. To reduce computational cost, we do not increase channel numbers as spatial sizes decrease.

Following the pyramid structure, a subsequent deformable alignment is cascaded to further refine the coarsely aligned features (the part with light purple background in Fig.~\ref{fig:pcd_align}). 
PCD module in such a coarse-to-fine manner improves the alignment to the sub-pixel accuracy. We demonstrate the effectiveness of PCD in Sec.~\ref{subsec:ablation_study}.
It is noteworthy that the PCD alignment module is jointly learned together with the whole framework, without additional supervision~\cite{tian2018tdan} or pretraining on other tasks like optical flow~\cite{xue2017video}. 

\begin{figure}[t]
	\vspace{-0.5cm}
	\begin{center}
		\includegraphics[width=0.9\linewidth]{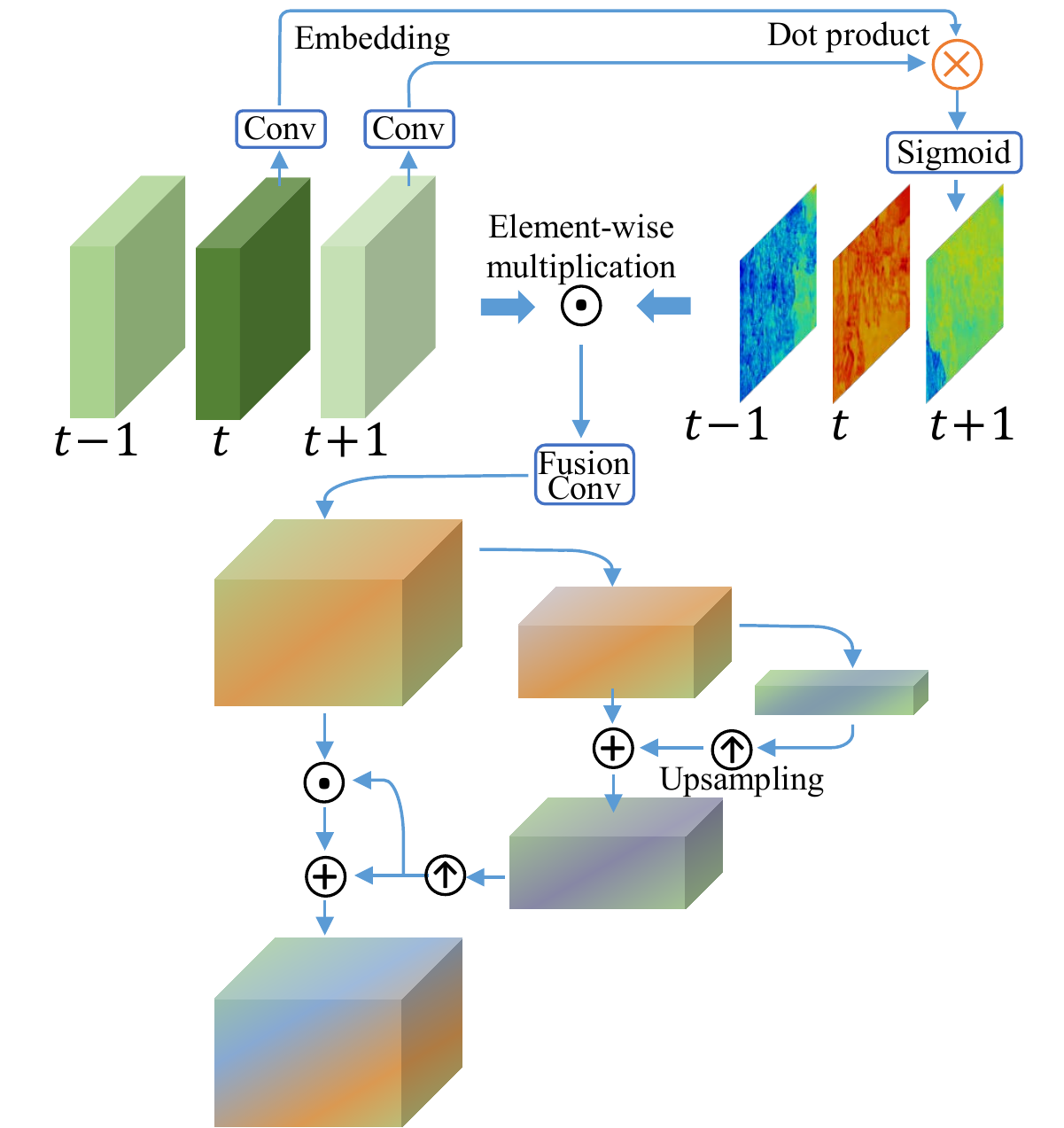}
		\vspace{-0.3cm}
		\caption{TSA fusion module with Temporal and Spatial Attention.}
		\label{fig:tsa_fusion}
		\vspace{-0.7cm}
	\end{center}
\end{figure}

\subsection{Fusion with Temporal and Spatial Attention}
\label{subsec:tsa}

Inter-frame temporal relation and intra-frame spatial relation are critical in fusion because 1) different neighboring frames are not equally informative due to occlusion, blurry regions and parallax problems; 2) misalignment and unalignment arising from the preceding alignment stage adversely affect the subsequent reconstruction performance. 
Therefore, dynamically aggregating neighboring frames in pixel-level is indispensable for effective and efficient fusion.
In order to address the above problems, we propose TSA fusion module to assign pixel-level aggregation weights on each frame. Specifically, we adopt temporal and spatial attentions during the fusion process, as shown in Fig.~\ref{fig:tsa_fusion}.

The goal of temporal attention is to compute frame similarity in an embedding space.
Intuitively, in an embedding space, a neighboring frame that is more similar to the reference one, should be paid more attention.
For each frame $i {\in} [{-}N{:}{+}N]$, the similarity distance $h$ can be calculated as:
\vspace{-0.1cm}
\begin{equation}
h({F_{t+i}^a}, {F_t^a}) = \text{sigmoid}(\ \theta({F_{t+i}^a})^T\phi({F_t^a})\ ),
\end{equation}
where $\theta({F_{t+i}^a})$ and $\phi({F_t^a})$ are two embeddings, which can be achieved with simple convolution filters. The sigmoid activation function is used to restrict the outputs in $[0,1]$, stabilizing gradient back-propagation.
Note that for each spatial location, the temporal attention is spatial-specific, \ie, the spatial size of $h({F_{t+i}^a}, {F_t^a})$ is the same as that of ${F_{t+i}^a}$.

The temporal attention maps are then multiplied in a pixel-wise manner to the original aligned features ${F_{t+i}^a}$. An extra fusion convolution layer is adopted to aggregate these attention-modulated features $\tilde{F}_{t+i}^a$:
\vspace{-0.1cm}
\begin{equation}
\tilde{F}_{t+i}^a = {F_{t+i}^a} \odot h({F_{t+i}^a}, {F_t^a}),
\vspace{-0.1cm}
\end{equation}
\begin{equation}
F_{\text{fusion}} = \text{Conv}(\ [\tilde{F}_{t-N}^a, \cdots, \tilde{F}_{t}^a, \cdots, \tilde{F}_{t+N}^a]\ ),
\end{equation}
where $\odot$ and $[\cdot, \cdot, \cdot]$ denote the element-wise multiplication and concatenation, respectively.

Spatial attention masks are then computed from the fused features. A pyramid design is employed to increase the attention receptive field.
After that, the fused features are modulated by the masks through element-wise multiplication and addition, similar to~\cite{wang2018recovering}. 
The effectiveness of TSA module is presented in Sec.~\ref{subsec:ablation_study}.

\subsection{Two-Stage Restoration}
\label{subsec:twostage}
\vspace{-0.2cm}
Though a single EDVR equipped with PCD alignment module and TSA fusion module could achieve state-of-the-art performance, it is observed that the restored images are not perfect, especially when the input frames are blurry or severely distorted. 
In such a harsh circumstance, motion compensation and detail aggregation are affected, resulting in inferior reconstruction performance. 

Intuitively, coarsely restored frames would greatly mitigates the pressure for alignment and fusion.
Thus, we employ a two-stage strategy to further boost the performance. Specifically, a similar but shallower EDVR network is cascaded to refine the output frames of the first stage. The benefits are two-fold: 1) it effectively removes the severe motion blur that cannot be handled in the preceding model, improving the restoration quality; 2) it alleviates the inconsistency among output frames.
The effectiveness of two-stage restoration is illustrated in Sec.~\ref{subsec:evaluation_reds}.

\section{Experiments} \label{sec:experiments}

\subsection{Training Datasets and Details}
\noindent\textbf{Training datasets}.
Previous studies on video processing~\cite{liu2014bayesian,jo2018deep,su2017deep} are usually developed or evaluated on private datasets.
The lack of standard and open video datasets restricts fair comparisons.
%
REDS~\cite{Nah_2019_CVPR_Workshops_REDS} is a newly proposed high-quality (720p) video dataset in the NTIRE19 Competition. 
REDS consists of 240 training clips, 30 validation clips and 30 testing clips (each with 100 consecutive frames).
During the competition, since the test ground truth is not available, we select four representative clips (with diverse scenes and motions) as our test set, denoted by \textit{REDS4}\footnote{Specifically, REDS4 contains the $000$, $011$, $015$ and $020$ clips.}. The remaining training and validation clips are re-grouped as our training dataset (a total of 266 clips).
To be consistent with our methods and process in the competition, we also adopt this configuration in this paper.

Vimeo-90K~\cite{xue2017video} is a widely used dataset for training, usually along with Vid4~\cite{liu2014bayesian} and Vimeo-90K testing dataset (denoted by Vimeo-90K-T) for evaluation. We observe dataset bias when the distribution of training sets deviates from that of testing sets. More details are presented in Sec.~\ref{subsec:ablation_study}.

\noindent\textbf{Training details}.
The PCD alignment module adopts five residual blocks (RB) to perform feature extraction. We use 40 RBs in the reconstruction module and 20 RBs for the second-stage model. The channel size in each residual block is set to $128$.
We use RGB patches of size $64{\times}64$ and $256{\times}256$ as inputs for video SR and deblurring tasks, respectively. Mini-batch size is set to 32.
The network takes five consecutive frames (\ie, N=2) as inputs unless otherwise specified.
We augment the training data with random horizontal flips and $90^{\circ}$ rotations. We only adopt Charbonnier penalty function~\cite{lai2017deep} as the final loss, defined by $\mathcal{L}{=}\sqrt{{\| \hat{O}_t{-}O_t\|}^2{+}\varepsilon^2}$, where $\varepsilon$ is set to $1{\times}10^{-3}$.

We train our model with Adam optimizer~\cite{kingma2014adam} by setting $\beta_1{=}0.9$ and $\beta_2{=}0.999$. The learning rate is initialized as $4{\times}10^{-4}$. We initialize deeper networks by parameters from shallower ones for faster convergence.
We implement our models with the PyTorch framework and train them using 8 NVIDIA Titan Xp GPUs.

\begin{table*}[!t]
	\small
	\vspace{-0.4cm}
	\caption{Quantitative comparison on \textbf{Vid4} for $4\times$ video SR. \redbf{Red} and \blueud{blue} indicates the best and the second best performance, respectively. Y or RGB denotes the evaluation on Y (luminance) or RGB channels. `*' means the values are taken from their publications.}
	\label{tab:sr_vid4}
	\vspace{-0.7cm}
	\begin{center}
		\tabcolsep=0.06cm
		\scalebox{0.93}{
			\begin{tabular}{l||c|c||c|c|c|c|c|c|c}
				\hline
				& Bicubic & RCAN~\cite{zhang2018image} & VESPCN*~\cite{caballero2017real} & SPMC~\cite{tao2017detail} & TOFlow~\cite{xue2017video} &FRVSR*~\cite{sajjadi2018frame} & DUF~\cite{jo2018deep} & RBPN*~\cite{haris2019recurrent} & \textbf{EDVR (Ours)} \\
				Clip Name & (1 Frame) & (1 Frame) & (3 Frames) & (3 Frames)  & (7 Frames)&(recurrent) & (7 Frames) & (7 Frames) & (7 Frames) \\ \hline
				Calendar (Y) & 20.39/0.5720 & 22.33/0.7254  & - & 22.16/0.7465 & 22.47/0.7318& - & \blueud{24.04}/\blueud{0.8110} & 23.99/0.807 & \redbf{24.05}/\redbf{0.8147} \\
				City (Y) & 25.16/0.6028 & 26.10/0.6960 & - & 27.00/0.7573  & 26.78/0.7403& - & \redbf{28.27}/\redbf{0.8313} & 27.73/0.803 & \blueud{28.00}/\blueud{0.8122} \\
				Foliage (Y) & 23.47/0.5666 & 24.74/0.6647  & - & 25.43/0.7208 & 25.27/0.7092& - & \redbf{26.41}/\redbf{0.7709} & 26.22/0.757 & \blueud{26.34}/\blueud{0.7635} \\
				Walk (Y) & 26.10/0.7974 & 28.65/0.8719  & - & 28.91/0.8761& 29.05/0.8790& -  & 30.60/\blueud{0.9141} & \blueud{30.70}/0.909 & \redbf{31.02}/\redbf{0.9152} \\ \hline
				Average (Y) & 23.78/0.6347 & 25.46/0.7395  & 25.35/0.7557 & 25.88/0.7752 & 25.89/0.7651& 26.69/0.822 & \blueud{27.33}/\redbf{0.8318} & 27.12/0.818 & \redbf{27.35}/\blueud{0.8264} \\ \hline
				Average (RGB) & 22.37/0.6098 & 24.02/0.7192  & -/- & 24.39/0.7534 & 24.41/0.7428& -/- & \blueud{25.79}/\redbf{0.8136} & -/- & \redbf{25.83}/\blueud{0.8077} \\ \hline
			\end{tabular}
		}
	\end{center}
	\vspace{-0.1cm}
\end{table*}
\begin{table*}[!t]
	\small
	\vspace{-0.4cm}
	\caption{Quantitative comparison on \textbf{Vimeo-90K-T} for $4\times$ video SR. `$\dagger$' means the values are taken from~\cite{xue2017video}. `*' means the values are taken from their publications.}
	\label{tab:sr_vimeo90k}
	\vspace{-0.7cm}
	\begin{center}
		\tabcolsep=0.06cm
		\scalebox{0.95}{
			\begin{tabular}{l||c|c||c|c|c|c|c|c}
				\hline
				& Bicubic & RCAN~\cite{zhang2018image} & DeepSR$^\dagger$~\cite{liao2015video} & BayesSR$^\dagger$~\cite{liu2014bayesian} & TOFlow~\cite{xue2017video} & DUF~\cite{jo2018deep} & RBPN*~\cite{haris2019recurrent} & \textbf{EDVR (Ours)} \\
				Test on & (1 Frame) & (1 Frame) & (7 Frames) & (7 Frames) & (7 Frames) & (7 Frames) & (7 Frames) & (7 Frames) \\ \hline
				RGB Channels & 29.79/0.8483 & 33.61/0.9101 & 25.55/0.8498 & 24.64/0.8205 & 33.08/0.9054 & \blueud{34.33}/\blueud{0.9227} & -/- & \redbf{35.79}/\redbf{0.9374} \\
				Y Channel & 31.32/0.8684 & 35.35/0.9251 & -/- & -/- & 34.83/0.9220 & 36.37/0.9387 & \blueud{37.07}/\blueud{0.9435} & \redbf{37.61}/\redbf{0.9489} \\ \hline
			\end{tabular}}
		\end{center}
		\vspace{-0.1cm}
	\end{table*}
\begin{table*}[!th]
	\small
	\vspace{-0.4cm}
	\caption{Quantitative comparison on \textbf{REDS4}. \textbf{Left}: $4\times$ Video SR (clean); \textbf{Right}: Video deblurring (clean). Test on RGB channels.}
	\label{tab:reds}
	\vspace{-0.7cm}
	\begin{center}
		\tabcolsep=0.01cm
		\scalebox{0.8}{
			\begin{subtable}[t]{0.5\textwidth}
				\hspace*{-2.3cm}
				\begin{tabular}{l||c|c|c|c|c}
					\hline
					Method & Clip\_000 & Clip\_011 & Clip\_015 & Clip\_020 & Average \\ \hline
					Bicubic & 24.55/0.6489 & 26.06/0.7261 & 28.52/0.8034 & 25.41/0.7386 & 26.14/0.7292 \\
					RCAN~\cite{zhang2018image} & 26.17/0.7371 & \blueud{29.34}/\blueud{0.8255} & \blueud{31.85}/\blueud{0.8881} & \blueud{27.74}/\blueud{0.8293} & \blueud{28.78}/0.8200 \\ \hline
					TOFlow~\cite{xue2017video} & 26.52/0.7540 & 27.80/0.7858 & 30.67/0.8609 & 26.92/0.7953 & 27.98/0.7990 \\
					DUF~\cite{jo2018deep} & \blueud{27.30}/\blueud{0.7937} & 28.38/0.8056 & 31.55/0.8846 & 27.30/0.8164 & 28.63/\blueud{0.8251} \\
					\textbf{EDVR (Ours)} & \redbf{28.01}/\redbf{0.8250} & \redbf{32.17}/\redbf{0.8864} & \redbf{34.06}/\redbf{0.9206} & \redbf{30.09}/\redbf{0.8881} & \redbf{31.09}/\redbf{0.8800} \\ \hline
				\end{tabular}
			\end{subtable}
			\begin{subtable}[t]{0.5\textwidth}
				\begin{tabular}{l||c|c|c|c|c}
					\hline
					Method & Clip\_000 & Clip\_011 & Clip\_015 & Clip\_020 & Average \\ \hline
					DeblurGAN~\cite{kupyn2018deblurgan} & 26.57/0.8597 & 22.37/0.6637 & 26.48/0.8258 & 20.93/0.6436 & 24.09/0.7482 \\
					DeepDeblur~\cite{nah2017deep} & 29.13/\blueud{0.9024} & 24.28/\blueud{0.7648} & 28.58/0.8822 & 22.66/0.6493 & 26.16/\blueud{0.8249} \\
					SRN-Deblur~\cite{tao2018srndeblur} & 28.95/0.8734 & \blueud{25.48}/0.7595 & 29.26/0.8706 & \blueud{24.21}/\blueud{0.7528} & \blueud{26.98}/0.8141 \\ \hline
					DBN~\cite{su2017deep} & \blueud{30.03}/0.9015 & 24.28/0.7331 & \blueud{29.40}/\blueud{0.8878} & 22.51/0.7039 & 26.55/0.8066 \\
					\textbf{EDVR (Ours)} & \redbf{36.66}/\redbf{0.9743} & \redbf{34.33}/\redbf{0.9393} & \redbf{36.09}/\redbf{0.9542} & \redbf{32.12}/\redbf{0.9269} & \redbf{34.80}/\redbf{0.9487} \\ \hline
				\end{tabular}
			\end{subtable}
		}
		
	\end{center}
\end{table*}

\begin{figure*}[!th]
	\vspace{-0.5cm}
	\begin{center}
		\includegraphics[width=\linewidth]{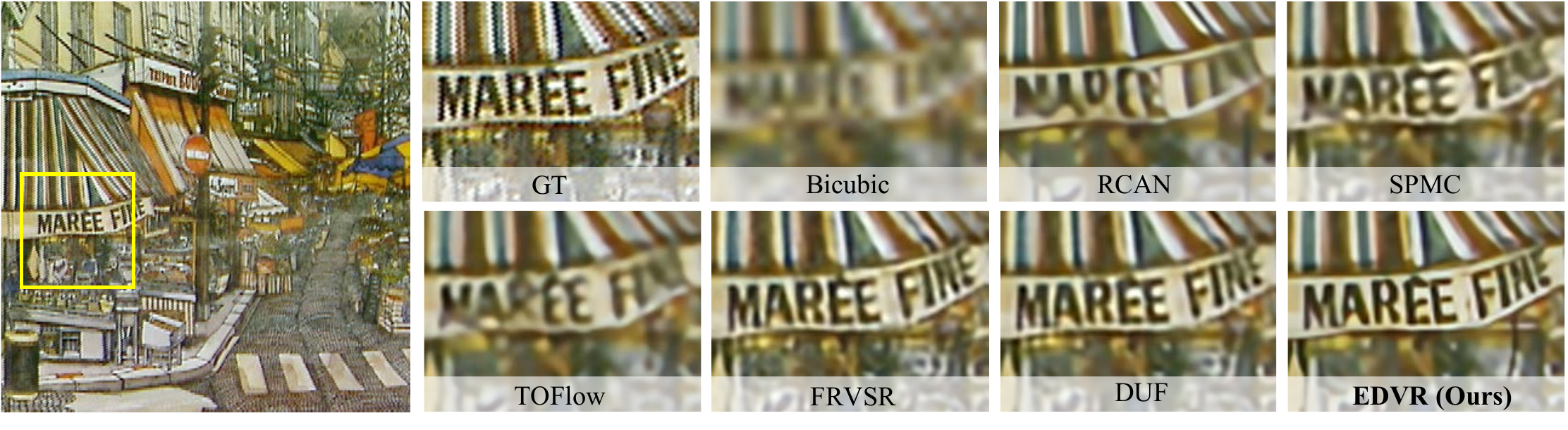}
		\vspace{-0.7cm}
		\caption{Qualitative comparison on the \textbf{Vid4} dataset for $4\times$ video SR. \textbf{Zoom in for best view.}}
		\label{fig:sr_vid4}
	\end{center}
	\vspace{-0.3cm}
\end{figure*}
\begin{figure*}[!th]
	\vspace{-0.35cm}
	\begin{center}
		\includegraphics[width=\linewidth]{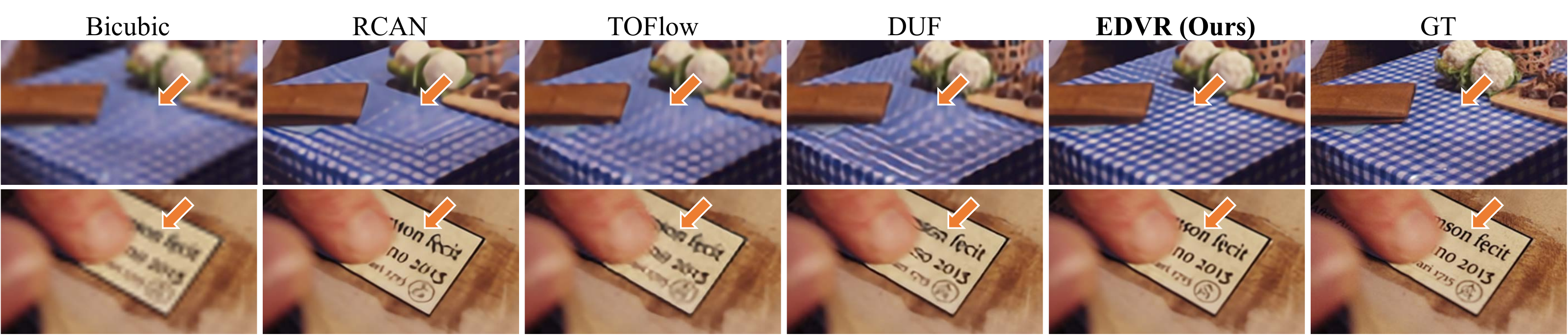}
		\vspace{-0.7cm}
		\caption{Qualitative comparison on the \textbf{Vimeo-90K-T} dataset for $4\times$ video SR. \textbf{Zoom in for best view.}}
		\label{fig:sr_vimeo90k}
	\end{center}
	\vspace{-0.35cm}
\end{figure*}
\begin{figure*}[!th]
	\vspace{-0.3cm}
	\begin{center}
		\includegraphics[width=\linewidth]{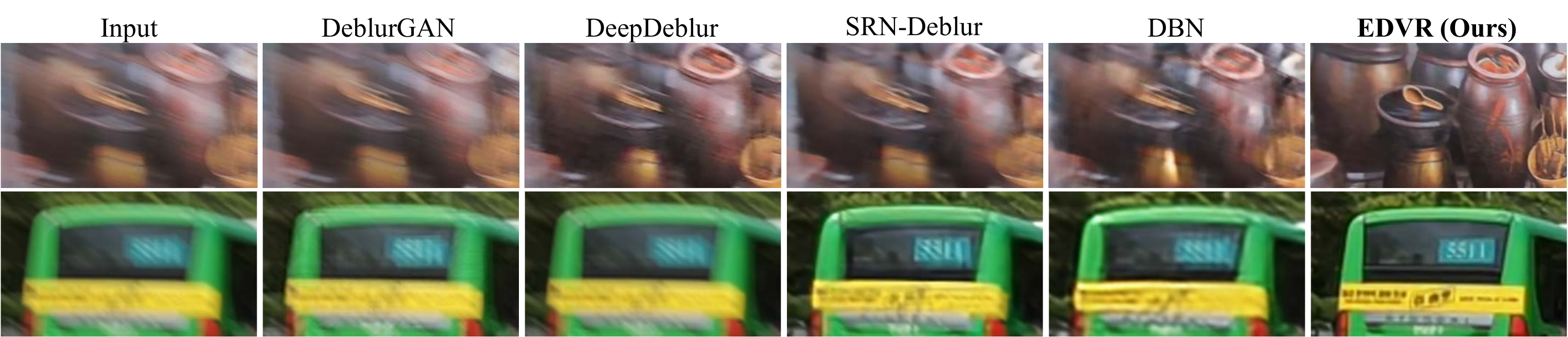}
		\vspace{-0.7cm}
		\caption{Qualitative comparison on the \textbf{REDS4} dataset for video deblurring. \textbf{Zoom in for best view.}}
		\label{fig:deblur:reds}
	\end{center}
	\vspace{-0.3cm}
\end{figure*}

\subsection{Comparisons with State-of-the-art Methods}
We compare our EDVR with several state-of-the-art methods on video SR and video deblurring respectively.
Two-stage and self-ensemble strategies~\cite{lim2017enhanced} are \textit{not} used.
In the evaluation, we include all the input frames and do not crop any border pixels except the DUF method~\cite{jo2018deep}. We crop eight pixels near image boundary for DUF due to its severe boundary effects.

\begin{table*}[!t]
	\small
	\vspace{-0.5cm}
	\caption{Ablations on:  \textbf{Left}: PCD and TSA modules (Experiments here adopt a smaller model with 10 RBs in the reconstruction module and the channel number is set to 64). FLOPs~\cite{molchanov2016pruning} are calculated on an image with the HR size of $1280\times720$. \textbf{Right}: the bias between the training and testing datasets.}
	\label{tab:ablations}
	\tabcolsep=0.1cm
	\vspace{-0.6cm}
	\begin{center}
		\scalebox{0.98}{
			\begin{subtable}[t]{0.45\textwidth}
				\hspace*{0.0cm}
				\begin{tabular}{l||c|c|c|c}
					\hline
					Model & Model 1 & Model 2 & Model 3 & Model 4 \\ \hline
					PCD? & \xmark\ (1 DConv) & \xmark\ (4 DConv)  & \cmark & \cmark \\
					TSA? & \xmark & \xmark & \xmark & \cmark \\ \hline
					PSNR & 29.78 & 29.98 & 30.39 & 30.53 \\
					FLOPs & 640.2G & 932.9G & 939.3G & 936.5G \\ \hline
				\end{tabular}
			\end{subtable}
			\begin{subtable}[t]{0.55\textwidth}
				\hspace*{0.1cm}
				\begin{tabular}{c||c||c|c}
					\hline
					\diagbox[trim=l]{Train}{Test}& REDS4 & Vid4~\cite{liu2014bayesian} & Vimeo90k~\cite{xue2017video} \\ \hline
					REDS (5 frames) & \textbf{31.09}/\textbf{0.8800} & 25.37/0.7956 & 34.33/0.9246 \\
					Vimeo-90K(7 frames) & 30.49/0.8700 & \textbf{25.83}/\textbf{0.8077} & \textbf{35.79}/\textbf{0.9374} \\ \hline
					$\Delta$ & 0.60/0.0100 & -0.46/-0.0121 & -1.46/-0.0128 \\ \hline
				\end{tabular}
			\end{subtable}
		}
	\end{center}
	\vspace{-0.1cm}
\end{table*}

\begin{figure*}[!th]
	\vspace{-0.4cm}
	\begin{center}
		\includegraphics[width=\linewidth]{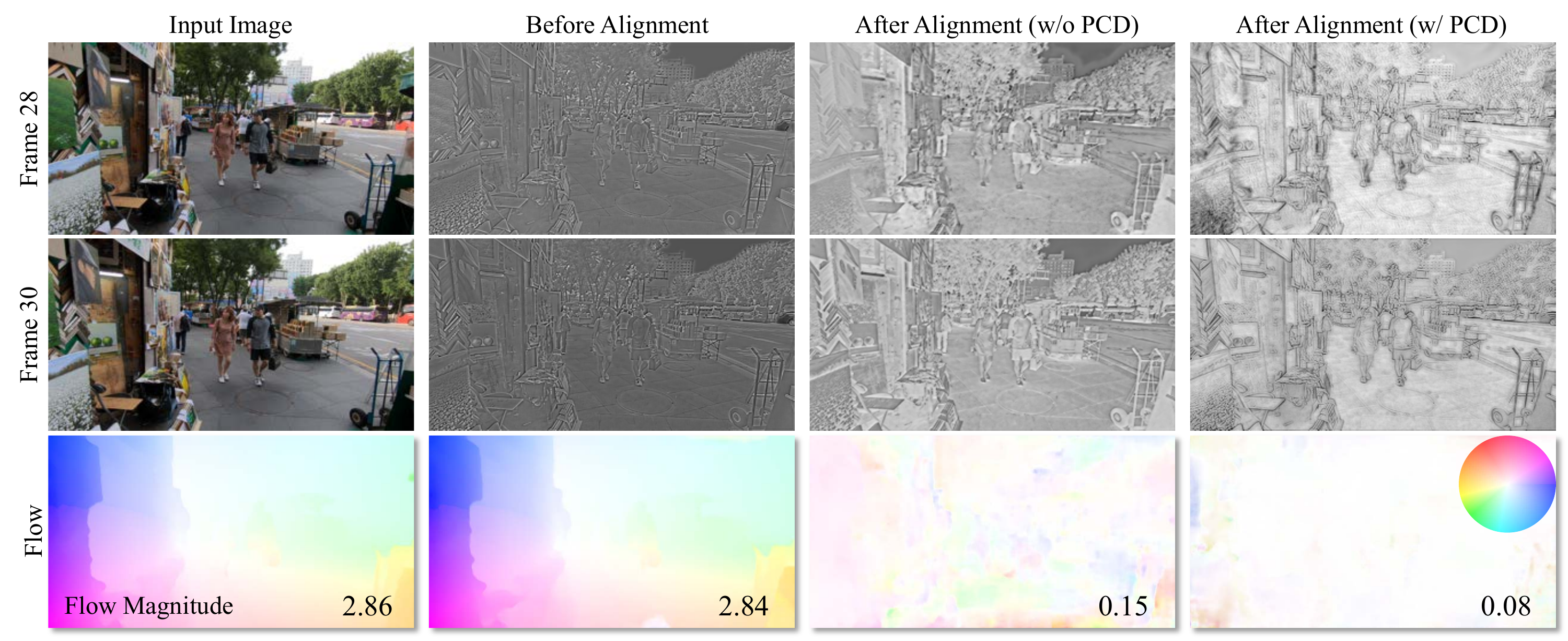}
		\vspace{-0.6cm}
		\caption{Ablation on the PCD alignment module. Compared with the results without PCD alignment, the flow of the PCD outputs is much smaller and cleaner, indicating that the PCD module can successfully handle large and complex motions. \textit{Flow field color coding scheme} is shown in the right. The direction and magnitude of the displacement vector are represented by hue and color intensity, respectively.}
		\label{fig:pcd_flow}
	\end{center}
	\vspace{-0.3cm}
\end{figure*}
\begin{figure*}[!th]
	\vspace{-0.4cm}
	\begin{center}
		\includegraphics[width=\linewidth]{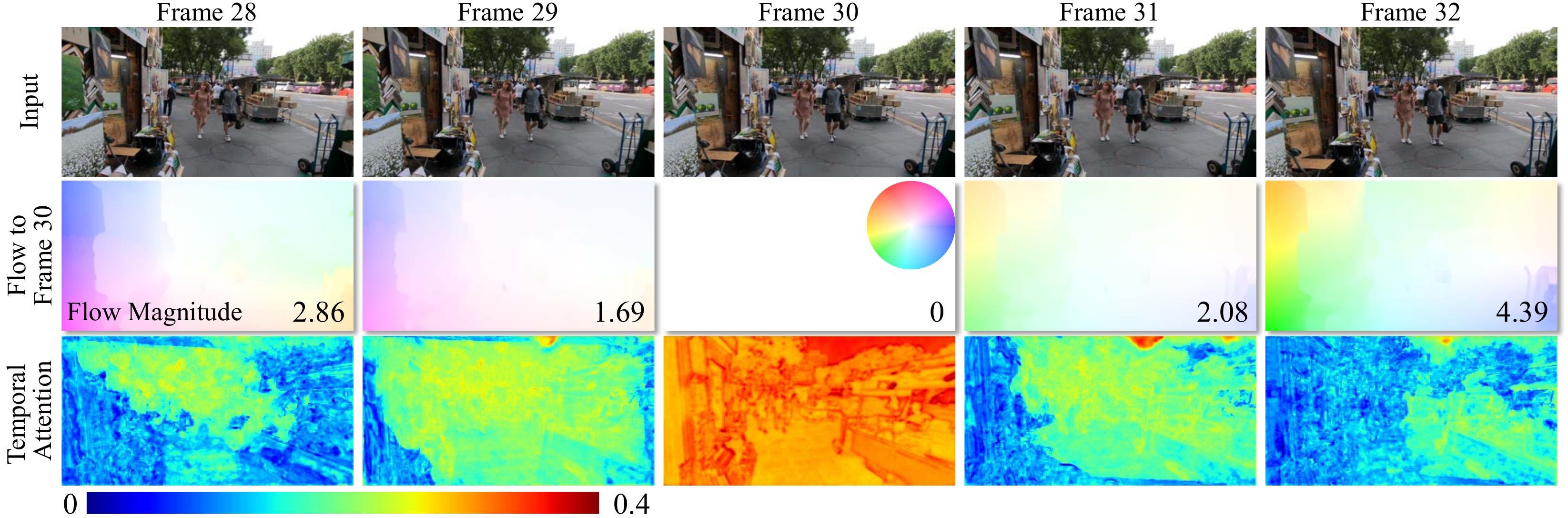}
		\vspace{-0.6cm}
		\caption{Ablation on the TSA fusion module. The frames and regions of lower flow magnitude tend to have more attention, indicating that the corresponding frames and regions are more informative.}
		\label{fig:tsa_flow}
	\end{center}
	\vspace{-0.9cm}
\end{figure*}

\noindent\textbf{Video Super-Resolution}.
We compare our EDVR method with nine algorithms: RCAN~\cite{zhang2018image}, DeepSR~\cite{liao2015video}, BayesSR~\cite{liu2014bayesian}, VESPCN~\cite{caballero2017real}, SPMC~\cite{tao2017detail}, TOFlow~\cite{xue2017video}, FRVSR~\cite{sajjadi2018frame}, DUF~\cite{jo2018deep} and RBPN~\cite{haris2019recurrent} on three testing datasets: Vid4~\cite{liu2014bayesian}, Vimeo-90K-T~\cite{xue2017video} and REDS4.
Most previous methods use different training sets and different down-sampling kernels, making the comparisons difficult.
Each testing dataset has different characteristics.
Vid4 is commonly used in video SR. The data has limited motion. Visual artifacts also exist on its ground-truth (GT) frames.
Vimeo-90K-T is a much larger dataset with various motions and diverse scenes. REDS4 consists of high-quality images but with larger and more complex motions.
We observe dataset bias when training and testing sets diverge a lot. Hence, we train our models on Vimeo-90K when evaluated on Vid4 and Vimeo-90K-T.

The quantitative results on Vid4, Vimeo-90K-T and REDS4 are shown in Table~\ref{tab:sr_vid4}, Table~\ref{tab:sr_vimeo90k} and Table~\ref{tab:reds} (Left), respectively. On Vid4, EDVR achieves comparable performance to DUF and outperforms other methods by a large margin. On Vimeo-90K-T and REDS, EDVR is significantly better than the state-of-the-art methods, including DUF and RBPN. 
Qualitative results on Vid4 and Vimeo-90K-T are presented in Fig.~\ref{fig:sr_vid4} and Fig.~\ref{fig:sr_vimeo90k}, respectively. On both datasets, EDVR recovers more accurate textures compared to existing methods, especially in the second image of Fig.~\ref{fig:sr_vimeo90k}, where the characters can be correctly identified only in the outputs of EDVR.

\noindent\textbf{Video Deblurring}.
We compare our EDVR method with four algorithms: DeepDeblur\cite{nah2017deep}, DeblurGAN~\cite{kupyn2018deblurgan}, SRN-Deblur~\cite{tao2018srndeblur} and DBN~\cite{su2017deep} on the REDS4 dataset.
Quantitative results are shown in Table~\ref{tab:reds} (Right). Our EDVR outperforms the state-of-the-art methods by a large margin. We attribute this to both the effectiveness of our method and the challenging REDS dataset that contains complex blurring.
Visual results are presented in Fig.~\ref{fig:deblur:reds}, while most methods are able to address small blurring, only EDVR can successfully recover clear details from extremely blurry images.

\begin{figure*}[!th]
	\vspace{-1cm}
	\begin{center}
		\includegraphics[width=1\linewidth]{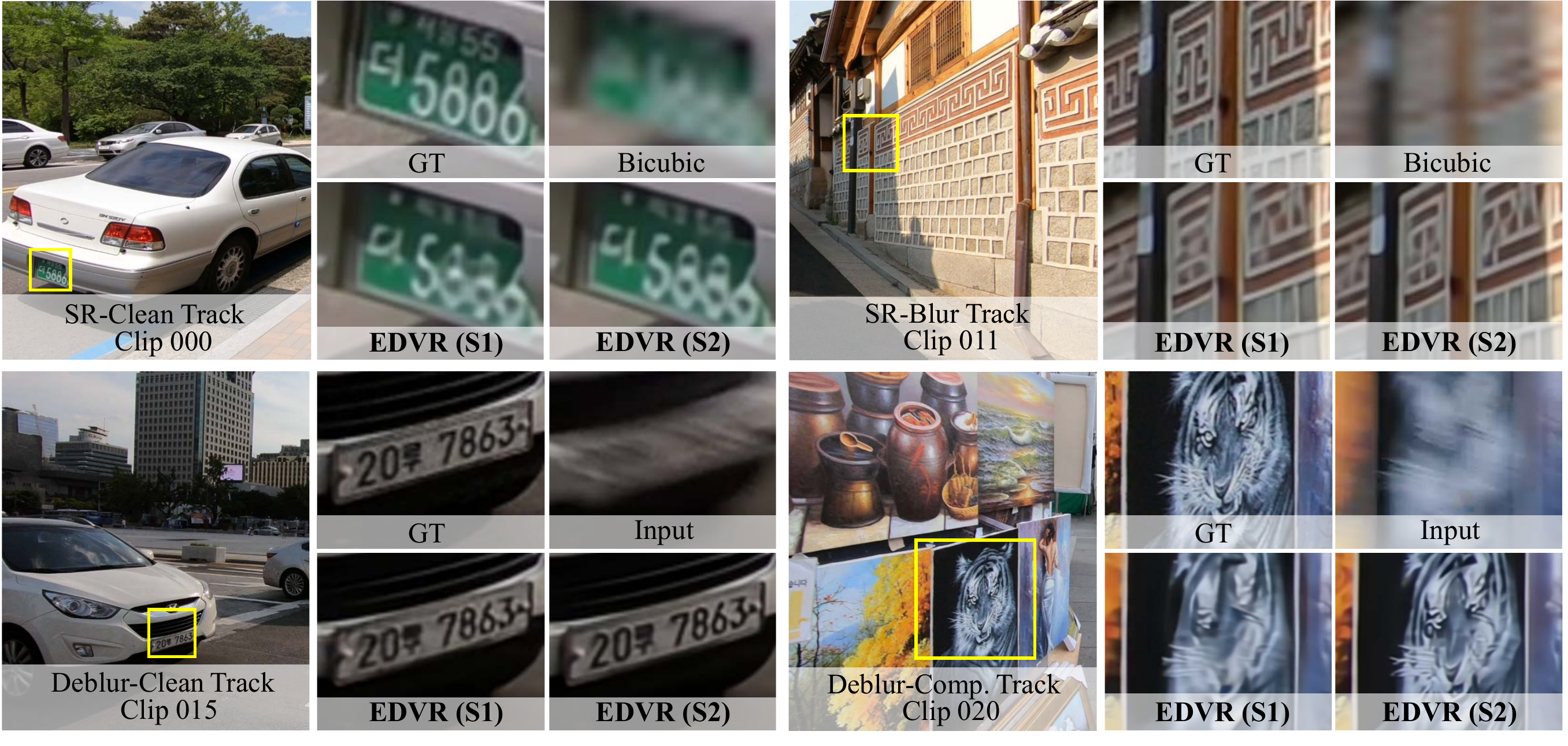}
		\vspace{-0.6cm}
		\caption{Qualitative results of our EDVR method on the four tracks in the NTIRE 2019 video restoration and enhancement challenges.}
		\label{fig:reds_4tracks}
		\vspace{-0.4cm}
	\end{center}
	\vspace{-0.4cm}
\end{figure*}

\subsection{Ablation Studies}
\label{subsec:ablation_study}
\vspace{-0.3cm}
\noindent \textbf{PCD Alignment Module.}
As shown in Table~\ref{tab:ablations} (Left), our baseline (Model 1) only adopts one deformable convolution for alignment. Model 2 follows the design of TDAN~\cite{tian2018tdan} to use four deformable convolutions for alignment, achieving an improvement of 0.2 dB. With our proposed PCD module, Model 3 is nearly 0.4 dB better than Model 2 with roughly the same computational cost, demonstrating the effectiveness of PCD alignment module. 
In Fig.~\ref{fig:pcd_flow}, we show representative features before and after different alignment modules, and depict the flow (derived by PWCNet~\cite{sun2018pwc}) between reference and neighboring features. Compared with the flow without PCD alignment, the flow of the PCD outputs is much smaller and cleaner, indicating that the PCD module can successfully handle large and complex motions.

\begin{table}[!t]
	\small
	\caption{Top5 methods in the NTIRE 2019 challenges on video restoration and enhancement. \redbf{Red} and \blueud{blue} indicates the best and the second best performance, respectively.}
	\label{tab:cometition_results}
	\tabcolsep=0.12cm
	\vspace{-0.65cm}
	\begin{center}
		\scalebox{0.88}{
			\hspace{-0.3cm}
			\begin{tabular}{l||cc|cc}
				\hline
				& \multicolumn{2}{c|}{SR} & \multicolumn{2}{c}{Deblur} \\
				Method & Clean & Blur & Clean & Compression \\ \hline
				\textbf{EDVR} (Ours) & \redbf{31.79}/\redbf{0.8962} & \redbf{30.17}/\redbf{0.8647} & \redbf{36.96}/\redbf{0.9657} & \redbf{31.69}/\redbf{0.8783} \\
				$2nd$ method & \blueud{31.13}/0.8811 & -/- & \blueud{35.71}/\blueud{0.9522} & \blueud{29.78}/0.8285 \\
				$3rd$ method & 31.00/\blueud{0.8822} & 27.71/0.8067 & 34.09/0.9361 & 29.63/\blueud{0.8261} \\
				$4th$ method & 30.97/0.8804 & 28.92/\blueud{0.8333} & 33.71/0.9363 & 29.19/0.8190 \\
				$5th$ method & 30.91/0.8782 & \blueud{28.98}/0.8307 & 33.46/0.9293 & 28.33/0.7976 \\ \hline
			\end{tabular}
		}
	\end{center}
	\vspace{-0.8cm}
\end{table}
\vspace{-0.2cm}
\noindent \textbf{TSA Attention Module.}
As shown in Table~\ref{tab:ablations} (Left), with the TSA attention module, Model 4 achieves 0.14 dB performance gain compared to Model 3 with similar computations. In Fig.~\ref{fig:tsa_flow}, we present the flow between the reference and neighboring frames, together with the temporal attention of each frame. It is observed that the frames and regions with lower flow magnitude tend to have higher attention, indicating that the smaller the motion is, the more informative the corresponding frames and regions are. 

\noindent \textbf{Dataset Bias.}
As shown in Table~\ref{tab:ablations} (Right), we conduct different settings of training and testing datasets for video super-resolution. The results show that there exists a large dataset bias. The performance decreases 0.5-1.5 dB when the distribution of training and testing data mismatch. We believe that the generalizability of video restoration methods is worth investigating.

\subsection{Evaluation on REDS Dataset}
\label{subsec:evaluation_reds}
We participated in all the four tracks in the NTIRE19 video restoration and enhancement challenges~\cite{Nah_2019_CVPR_Workshops_SR,Nah_2019_CVPR_Workshops_Deblur}. Quantitative results are presented in Table~\ref{tab:cometition_results}. Our EDVR wins the champions and outperforms the second place by a large margin in all tracks. 
In the competition, we adopt self-ensemble as~\cite{timofte2016seven,lim2017enhanced}. Specifically, during the test time, we flip and rotate the input image to generate \textit{four} augmented inputs for each sample. We then apply the EDVR method on each, reverse the transformation on the restored outputs and average for the final result. The two-stage restoration strategy as described in Sec.~\ref{subsec:twostage} is also used to boost the performance.
As shown in Table~\ref{tab:reds4}, we observe that the two-stage restoration largely improves the performance around 0.5 dB (EDVR(+) vs. EDVR-S2(+)). While the self-ensemble is helpful in the first stage (EDVR vs. EDVR+), it only brings marginal improvement in the second stage (EDVR-S2 vs. EDVR-S2+). 
Qualitative results are shown in Fig.~\ref{fig:reds_4tracks}. It is observed that the second stage helps recover clear details in challenging cases, \eg, the inputs are extremely blurry.
%

\begin{table}[t]
	\small
	\caption{Evaluation on REDS4 for all the four competition tracks. `+' and `-S2' denote the self-ensemble strategy and two-stage restoration strategy, respectively.}
	\label{tab:reds4}
	\tabcolsep=0.1cm
	\vspace{-0.65cm}
	\begin{center}
		\scalebox{0.9}{
			\tabcolsep=0.06cm
			\hspace{-0.3cm}
			\begin{tabular}{cc|c|c||c|c}
				\hline
				\multicolumn{2}{c|}{Track} & EDVR & EDVR-S2 & EDVR+ & EDVR-S2+ \\ \hline
				\multirow{2}{*}{SR} & Clean & 31.09/0.8800 & 31.54/0.8888 & 31.23/0.8818 & 31.56/0.8891 \\
				& Blur & 28.88/0.8361 & 29.41/0.8503 & 29.14/0.8403 & 29.49/0.8515 \\ \hline
				\multirow{2}{*}{Deblur} & Clean & 34.80/0.9487 & 36.37/0.9632 & 35.27/0.9526 & 36.49/0.9639 \\
				& Comp. & 30.24/0.8567 & 31.00/0.8734 & 30.46/0.8599 & 31.06/0.8741 \\ \hline
			\end{tabular}
		}
	\end{center}
	\vspace{-0.8cm}
\end{table}

\section{Conclusion}
We have introduced our winning approach in the NTIRE 2019 video restoration and enhancement challenges. To handle the challenging benchmark released in the competition, we propose EDVR, a unified framework with unique designs to achieve good alignment and fusion quality in various video restoration tasks.
Thanks to the PCD alignment module and TSA fusion module, EDVR not only wins all four tracks in the NTIRE19 Challenges but also demonstrates superior performance to existing methods on several benchmarks of video super-resolution and deblurring.

\vspace{0.2cm}
\normalsize{
	\noindent\textbf{Acknowledgement}.
	We thank Yapeng Tian for providing the core codes of TDAN~\cite{tian2018tdan}.
	This work is supported by SenseTime Group Limited, Joint Lab of CAS-HK, the General Research Fund sponsored by the Research Grants Council of the Hong Kong SAR (CUHK 14241716, 14224316. 14209217), and Singapore MOE AcRF Tier 1 (M4012082.020).
}
\clearpage
\clearpage
{\small
\bibliographystyle{ieee}
\bibliography{short,bib}
}

\end{document}